\documentclass[letterpaper, 10 pt, conference]{ieeeconf}  % Comment this line out if you need a4paper

\IEEEoverridecommandlockouts                              % This command is only needed if 
\usepackage[table]{xcolor}
\usepackage{graphics} % for pdf, bitmapped graphics files
\usepackage{epsfig} % for postscript graphics files
\usepackage{mathptmx} % assumes new font selection scheme installed
\usepackage{times} % assumes new font selection scheme installed
\usepackage{amsmath} % assumes amsmath package installed
\usepackage{amssymb}  % assumes amsmath package installed
\usepackage{xcolor}
\usepackage[colorlinks=true, linkcolor=blue, anchorcolor=blue, citecolor=blue, filecolor=blue, menucolor=blue, urlcolor=blue]{hyperref}
\usepackage{siunitx}
\usepackage{seqsplit}
\usepackage{subcaption}
\usepackage{multirow}
\usepackage{amsfonts}
\usepackage{bm}
\usepackage{placeins}

\DeclareMathAlphabet{\mathcal}{OMS}{cmsy}{m}{n}

\DeclareMathOperator*{\maxval}{max}

\title{\LARGE \bf
Robust Monocular Edge Visual Odometry through Coarse-to-Fine Data Association
}

\author{Xiaolong Wu, Patricio Vela, and C\'edric Pradalier% <-this % stops a space
}

\begin{document}

\maketitle
\thispagestyle{empty}
\pagestyle{empty}

%%%%%%%%%%%%%%%%%%%%%%%%%%%%%%%%%%%%%%%%%%%%%%%%%%%%%%%%%%%%%%%%%%%%%%%%%%%%%%%%
\begin{abstract}

In this work, we propose a monocular visual odometry framework, which allows exploiting the best attributes of edge feature for illumination-robust camera tracking, while at the same time ameliorating the performance degradation of edge mapping. In the front-end, an ICP-based edge registration can provide robust motion estimation and coarse data association under lighting changes. In the back-end, a novel edge-guided data association pipeline searches for the best photometrically matched points along geometrically possible edges through template matching, so that the matches can be further refined in later bundle adjustment. The core of our proposed data association strategy lies in a point-to-edge geometric uncertainty analysis, which analytically derives (1) the probabilistic search length formula that significantly reduces the search space for system speed-up and (2) the geometrical confidence metric for mapping degradation detection based on the predicted depth uncertainty. Moreover, match confidence based patch size adaption strategy is integrated into our pipeline, connecting with other components, to reduce the matching ambiguity. We present extensive analysis and evaluation of our proposed system on synthetic and real-world benchmark datasets under the influence of illumination changes and large camera motions, where our proposed system outperforms current state-of-art algorithms. 

\end{abstract}

%%%%%%%%%%%%%%%%%%%%%%%%%%%%%%%%%%%%%%%%%%%%%%%%%%%%%%%%%%%%%%%%%%%%%%%%%%%%%%%%
\section{INTRODUCTION}
\label{sec:introduction}
In recent decades, monocular Visual Odometry (VO) systems have shown their full potential to assist various outdoor robotic applications. Among these algorithms, indirect methods \cite{mur2017orb} are the \textit{de facto} standards due to the robustness of visual features against both photometric noise and lens distortion. However, recent works have shown that direct methods \cite{engel2017direct} could provide more robust motion estimation contributed by its more complete usage of information contained in the image.

\begin{figure}[t] 
  \centering
  \includegraphics[width=\linewidth]{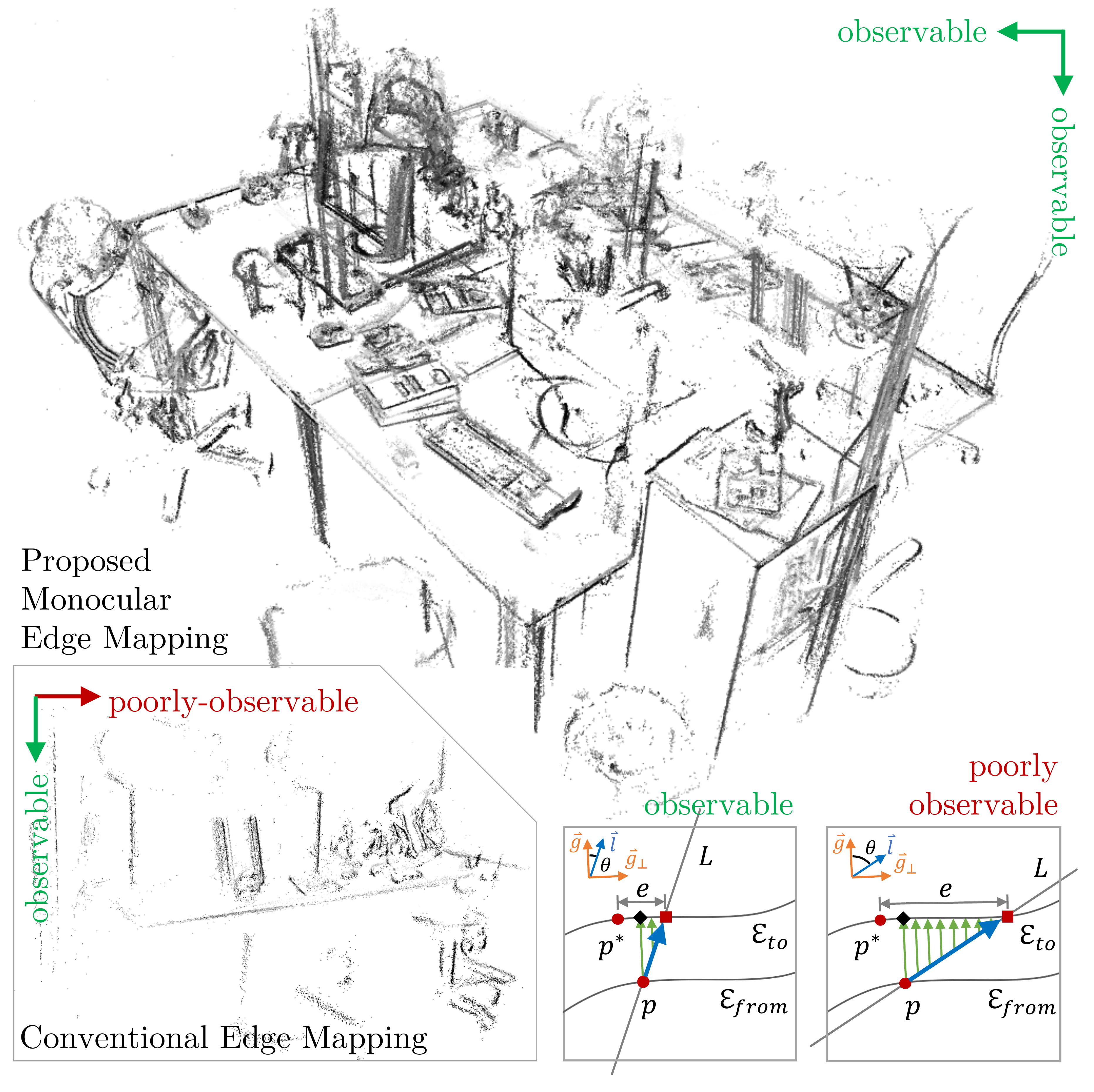}
    \caption{ \small \textbf{Our proposed Monocular Edge VO} (top) is capable of using edge feature and image gradient to overcome the partial observability issue of pure Edge Mapping (bottom-left). Pure edge mapping minimize point-to-tangent error ({\color{green!50!black} $\uparrow$}) that results in more erroneous matches ({\color{red} $\blacksquare$}) under poorly-observable direction (bottom-right) than observable direction (bottom-middle). }
\label{fig:frontpage}
\end{figure}

Edge alignment can be seen as a crossover of indirect and direct principles. Specifically, edges are geometric features extracted from raw images, but edge registration is performed using iterative-closest-point (ICP) based direct alignment \cite{kneip2015sdicp} \cite{zhou2017semi} \cite{zhou2018canny}. The first edge VO \cite{jose2015realtime} was developed to align edges by searching for the closest counterpart along the normal direction. Later, efficiency improvements for 2D-3D registration based on the Distance Transform (DT) \cite{kneip2015sdicp} improved the real-time properties of motion estimation. The optimizability of this formula is further improved by substituting DT with Approximate Nearest Neighbor Fields (ANNFs) \cite{zhou2017semi} or Oriented Nearest Neighbor Fields (ONNFs) \cite{zhou2018canny}, which have demonstrated strong performances for RGB-D sensors. However, it is still tricky to implement on monocular VO systems since the binary edge features are not self-distinguishable that makes the depth partially observable under the ICP-based mapping framework. Conventional depth estimation algorithms that search for best matches along epistolary lines may result in unreliable correspondences and lead to unstable and error-prone depth estimates in Fig. \ref{fig:frontpage}. 

Considering the partial observability issue of edge mapping, researchers have proposed a sophisticated group matching strategy \cite{tarrio2019se} to realize geometrically consistent matches, but fail to provide any theoretical guarantees about its correctness. Using optical flow technique \cite{bouguet2001pyramidal} to find the matches through photometric minimization makes the depth fully observable in the back-end \cite{maity2017edge}. However, the optical flow relies heavily on brightness constancy assumption, limiting its use for outdoor operations. 

As an alternative, illumination-robust tracking algorithms based on the Lucas-Kanade method \cite{baker2004lucas} have also been studied extensively \cite{park2017illumination}. An analysis suggests that gradient \cite{dai2017bundlefusion} and census transform \cite{alismail2016direct} approaches show state-of-the-art tracking accuracy, but considerably compromise the convergence basin. The reduced convergence radius arises from a flatter cost functions, which increases the sensitivity to perception noise and introduces artificial local optima from the noise..

One way to improve the illumination-robust template matching is to utilize confidence measures \cite{hu2012quantitative} to assess the correctness of the hypothesis, a technique widely used in the field of stereo vision for error detection \cite{park2015leveraging}. Among the various metrics, Attainable Maximum Likelihood (AML) \cite{merrell2007real} shows high performance for multi-view stereo matching, which is readily incorporated into our system for match ambiguity detection. 

Inspired by semi-direct methods \cite{forster2014svo} \cite{gao2018ldso} integrating the complementary strengths of indirect and direct formulations for superior performance, we propose a monocular edge VO framework in Fig. \ref{fig:system} integrating the illumination-robustness of edge features, the informativeness of photometric matching, and the efficiency of pose-graph optimization to solve the issues mentioned above. Our proposed framework inherently conforms to a coarse-to-fine data association structure, which iteratively refines the edge point correspondences by exploiting geometric and photometric information. The main contributions of this work ares:

\begin{itemize}
  \item A monocular edge VO framework, comprised of ICP-based edge alignment, edge-guided data association, and local BA, which is capable of performing illumination-robust camera tracking and scene reconstruction without incurring edge mapping degradation.  
  \item An edge-guided data association pipeline incorporating probabilistic search length approximation, image-gradient-based template matching, match-confidence-based patch size adaption, and depth-confidence-based match conditioning.
  \item A point-to-edge geometric uncertainty analysis that analytically derives a probabilistic search length formula and a depth confidence measure that improves the efficiency and accuracy of our proposed system.
\end{itemize}  

\section{SYSTEM OVERVIEW}
\label{sec:systemoverview}
\begin{figure}[t] 
  \centering
  \includegraphics[width=\linewidth]{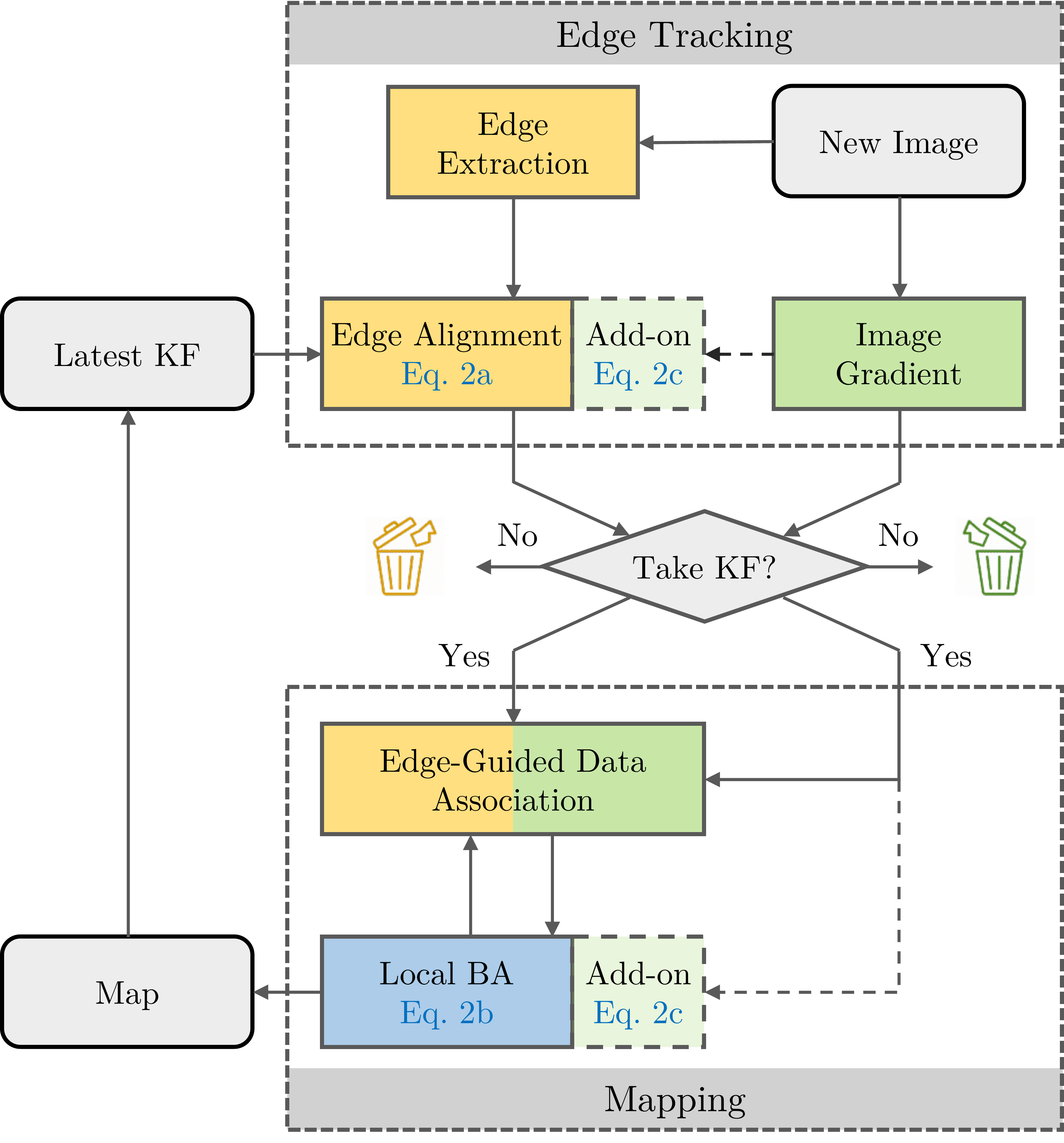}
    \caption{ \small \textbf{Our proposed monocular Edge VO framework} is a KeyFrame (KF) based monocular VO framework, which can be generally divided into edge alignment, edge-guided data association, and local BA.}
\label{fig:system}
\end{figure}

Fig. \ref{fig:system} illustrates our proposed monocular edge VO framework, which consists of two parallel threads named edge tracking and mapping. Our proposed edge tracking threads aim to coarsely estimate the camera motion and edge point association from the current frame to the latest keyframe, which is achieved through the minimization of point-to-tangent error in Eq. (\ref{eq:edgeerror}) using ICP pipeline. As soon as a new keyframe is created, the edge-guided data association module in Sec. \ref{sec:dataassociation} refines the correspondences by incorporating illumination-robust photometric information. Finally, the local BA jointly optimizes the camera poses and scene structure using resultant matches in Eq. (\ref{eq:reprojectionerror}) within a sliding window of keyframes. We also provide the option to involve the image-gradient-based costs Eq. (\ref{eq:photometricerror}) into both edge tracking and local BA layers, especially for the applications that hold the special emphasis on accuracy but real-time performance,

Note that both the edge tracking and local BA are well-studied problems, therefore we choose the state-of-the-art implementations for our system design. Our edge alignment front-end implements the ANNFs \cite{zhou2017semi}  \cite{zhou2018canny} with a pyramidal coarse-to-fine scheme for point-to-tangent registration in Eq. (\ref{eq:edgeerror}), which approximates the nearest neighbors as temporal correspondences for ICP-based optimization. Our local bundle adjustment algorithm performs a joint optimization of camera poses and scene structure altogether in Eq. (\ref{eq:reprojectionerror}), which can be achieved through pose-graph optimization \cite{grisetti2011g2o} \cite{dellaert2017gtsam} or a customized solver \cite{engel2017direct}. 

\begin{figure}[t] 
  \centering
  \includegraphics[width=\linewidth]{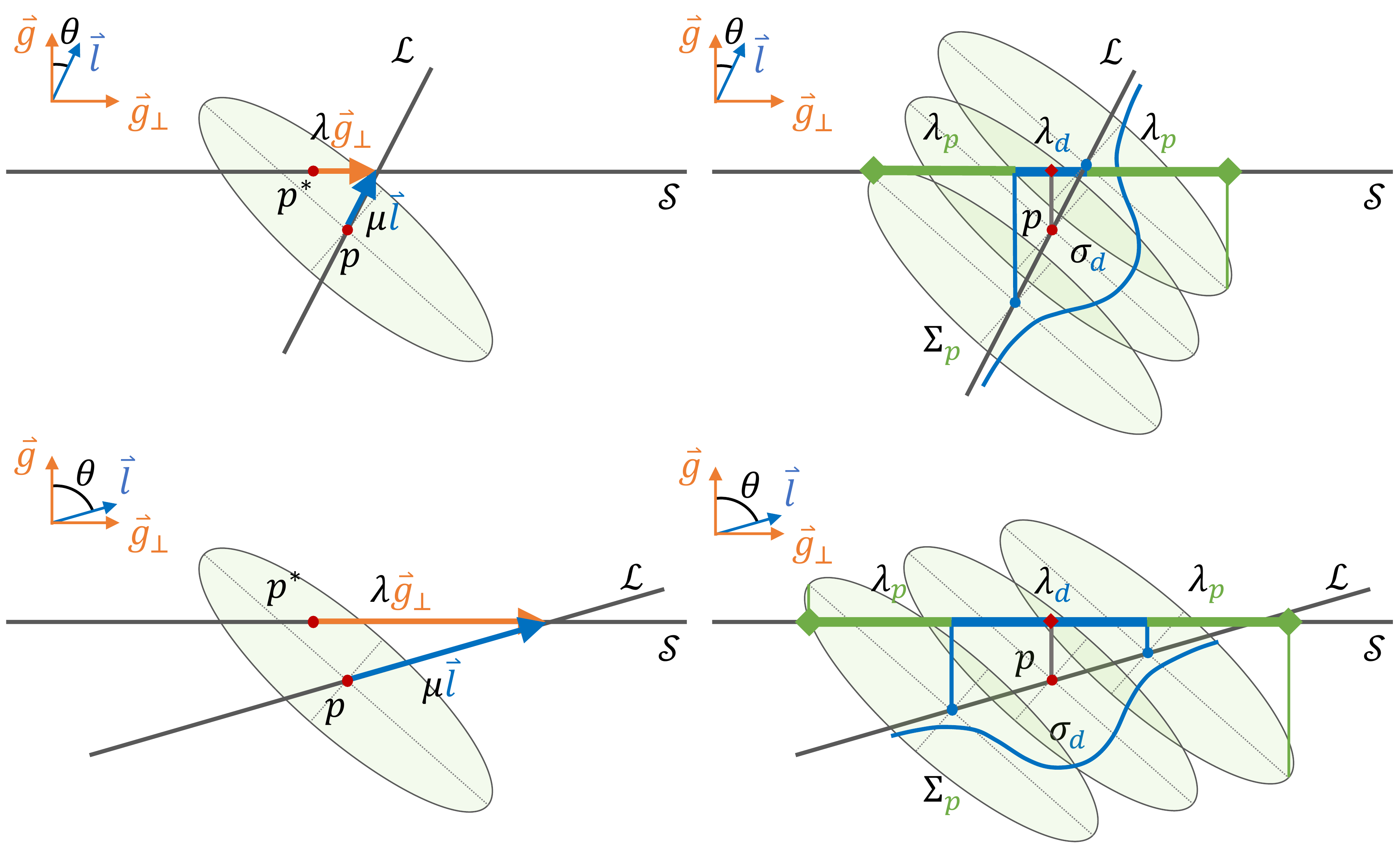}
    \caption{ \small \textbf{The Point-to-Edge Uncertainty Analysis} for observable (top: $\theta$ is small) and poorly-observable (bottom: $\theta$ is large) cases. The geometric relationship described in Eq. (\ref{eq:intersection}) (left) and the geometric interpretation of our derived edge search length in Eq. (\ref{eq:searchradius}) (right) are presented. }
\label{fig:searcharea}
\end{figure}
 
\section{PROBLEM FORMULATION}
\label{sec:formulation}
Consider an acquired image in the reference frame $I_{r}: \Omega \rightarrow \mathbb{R}$, where $\Omega \subset \mathbb{R}^2$ is the image domain. A 3D scene point $\mathbf{P} = (x, y, z)^{T}$ is parameterized by its inverse depth $d = z^{-1}$. Each pixel $\mathbf{p} = (u,v)^T \in \Omega$ can be back-projected into the 3D world using the back-projection function $\mathbf{P} = \pi^{-1}(\mathbf{p})$, and inversely using projective warp function $\mathbf{p} = \pi(\mathbf{P})$. A 3D rigid body transformation $\mathbf{G} \in SE(3)$ can be generally divided into a rotation $\mathbf{R} \in SO(3)$ and translation $\mathbf{t} \in \mathbb{R}^3$. In the optimization framework, $\mathbf{G}$ is represented as its corresponding Lie Group component $\mathbf{\xi} \in \mathfrak{se}(3)$, where this element can be mapped to $\mathbf{G} \in SE(3)$ through the exponential mapping described in \cite{blanco2010tutorial}.

Given arbitrary pixel selector $\mathcal{S}(\cdot)$, the group of edge pixels $\mathcal{S}_r = \left \{ \mathbf{p}_r \right \}$  are subsequently projected to current frame $k$ as:
\begin{equation} \label{eq:pointproj}
 \mathbf{p}_{kr} =  \pi ( \mathbf{R}_{kr}\pi^{-1}(\mathbf{p}_r,d_r)+\mathbf{t}_{kr} ) 
\end{equation} 
The point-to-tangent edge alignment error $E_{kr}^{\mathcal{E}}$, reprojection error $E_{kr}^{\mathcal{R}}$, and generalized photometric error $E_{kr}^{\mathcal{P}}$ from the reference frame to frame $k$ can be generally written as:
\begin{subequations}
\begin{align} 
E_{kr}^{\mathcal{E}} &:=  \sum_{\mathbf{p}_{r} \in \mathcal{S}_r^{\mathcal{E}} } w_{\mathbf{p}_r}^{\mathcal{E}} \Vert  \mathbf{g}^T( \mathbf{p}_{kr} - \mathit{n}(\mathbf{p}_{kr}) )  \Vert_{\gamma}  \label{eq:edgeerror} \\
E_{kr}^{\mathcal{R}} &:=  \sum_{\mathbf{p}_{r} \in \mathcal{S}_r^{\mathcal{R}} } w_{\mathbf{p}_r}^{\mathcal{R}} \Vert \mathbf{p}_{kr} - \mathbf{p}_{k} \Vert_{\gamma}  \label{eq:reprojectionerror} \\
E_{kr}^{\mathcal{P}} &:= \sum_{\mathbf{p}_r \in \mathcal{S}_r^{\mathcal{P}}} w_{\mathbf{p}_r}^{\mathcal{P}} \Vert F_{k} ( \mathbf{p}_{kr}) - F_{r}(\mathbf{p}_r) \Vert_{\gamma} \label{eq:photometricerror} 
\end{align}
\end{subequations}
where $w_{\mathbf{p}_r}$ is the weight assigned for each selected pixel from $\mathcal{S}_r$ in the reference frame, and $\Vert \cdot \Vert_{\gamma}$ is the Huber norm. Specifically for each formulation, $\mathit{n}(\cdot)$ represents the nearest neighbor edge pixel in current frame $k$ using the Euclidean distance metric, $\mathbf{g}$ is the abbreviation of $\mathbf{g}(\mathit{n}(\mathbf{p}_{kr}))$ representing the gradient direction vector of the temporal match of a given projected pixel $\mathbf{p}_{kr}$, $\left \{ \mathbf{p}_{r},\mathbf{p}_{k} \right \}$ is a pair of matched points between the reference image and image $k$, and $F(\cdot)$ represents any representation calculated from image $I$, such as intensity or gradient.

\section{EDGE-GUIDED DATA ASSOCIATION}
\label{sec:dataassociation}
This section presents a point-to-edge uncertainty analysis in Sec. \ref{ssec:uncertaintyanalysis}. The analysis informs out edge-guided data association pipeline in Sec. \ref{ssec:dapipeline}, which further refines the edge-point correspondences using geometric relationships and photometric information.

\subsection{Point-to-Edge Uncertainty Analysis}
\label{ssec:uncertaintyanalysis}
Point-to-edge analysis is carried out to realize the potential search length, that is, the error variance of the search radius along the edge direction caused by tracking error on $\xi$ and inverse depth error on $d$. To simplify the analysis, we make two assumptions: (1) the edge is locally linear so that its gradient direction is locally constant, and (2) the rotation error on $\xi$ plays a minor role in the epipolar line direction estimation. After such simplifications, the search line along edge direction $\mathcal{S}$ and epipolar line $\mathcal{L}$ can be approximated as:  
\begin{subequations}
\begin{gather} 
\mathcal{S} := \left \{ \mathbf{p}^* +  \lambda \mathbf{g}_{\perp} \right \} \ \ \ 
\mathcal{L} := \left \{ \mathbf{p} +  \mu \mathbf{l} \right \} \label{eq:searchradius} \\
 \mathbf{p}^* +  \lambda \mathbf{g}_{\perp} = \mathbf{p} +  \mu \mathbf{l} \label{eq:intersection}
\end{gather}
\end{subequations}
where $\mathbf{p}$ represents the reprojected edge point from the reference frame to the current frame, and $\mathbf{p}^{*}$ denotes its correspondence lying on the locally linear region of target edge. $\mathbf{g}_{\perp}$ and $\mathbf{l}$ are the normalized perpendicular epipolar line and image gradient directions, while $\lambda$ and $\mu$ are their distance factors. $\theta$ represents the angle between the normalized image gradient $\mathbf{g}$ and the epipolar line $\mathbf{l}$ vectors, which describe the angular relationship between $\mathcal{S}$ and $\mathcal{L}$. The described geometric relationship is illustrated in Fig. \ref{fig:searcharea}. 

\subsubsection{Probabilistic search length}
\label{ssec:searchlength}
Here, we derive the formula for the search length factor $\lambda$ and its variance $\sigma_{\lambda}^2$ w.r.t. the $\mu$ and $\mathbf{p}$. For simplicity, we assume the uncertainties of $\mu$ and $\mathbf{p}$ are independent, so that the derived search length function $\lambda$ and its variance $\sigma_{\lambda}^2$ can be expressed as:
\begin{subequations}
\begin{gather}
 \lambda( \mathbf{p},\mu) = \langle \mathbf{p}^* - \mathbf{p},  \mathbf{g}_{\perp} \rangle + \mu \langle \mathbf{l},\mathbf{g}_{\perp} \rangle = e_{p \perp g} + \mu sin \theta  \label{eq:radiusequation} \\
\sigma_{\lambda}^2( \mathbf{p},\mu) = \mathbf{J}_{p} \Sigma_{p} \mathbf{J}_{p}^T +  \mathbf{J}_{\mu} \sigma_{\mu}^2 \mathbf{J}_{ \mu}^T=  \sigma_{p \perp g}^2 + \sigma_{\mu}^2 \sin^2 \theta \label{eq:rediusuncertainty}
\end{gather}
\end{subequations}
where $\mathbf{J}_{p}$ and $\mathbf{J}_{\mu}$ are the Jacobians of Eq. (\ref{eq:radiusequation}) given the statistics of $\mathbf{p}$ and $\mu$. $e_{p \perp g}$ is the reprojection error component perpendicular to the edge normal direction $\mathbf{g}$. $\Sigma_{\mathbf{p}}$ and $\sigma_{\mu}^2$ denote the (co-)variances of reprojected point and depth disparity. For fast calculation, the upper bound of variance can be estimated using the inequality relationship as:
\begin{equation} \label{eq:uncertaintyseperation}
\sigma_{\lambda} =\Vert \sigma_{p \perp g}^2 + \sigma_{\mu}^2 \sin^2 \theta \Vert^{\frac{1}{2}} \leq \sigma_{p \perp g} + \sigma_{\mu} | \sin \theta |
\end{equation}
Setting the center of search to be the temporally estimated edge point correspondence $\mathit{n} \left( \mathbf{p} \right)$, the search radius $\lambda_{1/2}$ can be expressed as:
\begin{equation} \label{eq:searchregion}
\lambda_{1/2} =  k_p \sigma_{p \perp g} + k_{\mu} \sigma_{\mu} | \sin \theta | 
\end{equation}
where $k_p$ and $k_{\mu}$ denote the gains to compensate the potential shrinkage of search length due to our approximations. 

The point reprojection covariance $\Sigma_{p}$ can be readily calculated from edge alignment front-end using conventional uncertainty propagation, which can be decomposed into a group of more compact representations, the eigenvalue $\sigma$ and eigenvector $\mathbf{v}$, through eigenvalue decomposition as follows:
\begin{equation} \label{eq:poseuncertainty}
\Sigma_{p} =  \mathbf{J}_{\xi}^p (\sum_{p} {\mathbf{J}_{\xi}^p}^T \mathbf{J}_{\xi}^p) {\mathbf{J}_{\xi}^p}^T \sigma_{r}^2 =
\begin{bmatrix}
\mathbf{v}_1 \mathbf{v}_2 
\end{bmatrix}
\begin{bmatrix}
\sigma_1^2 & \\
 & \sigma_2^2 
\end{bmatrix}
\begin{bmatrix}
\mathbf{v}_1^T \\
\mathbf{v}_2^T 
\end{bmatrix}
\end{equation}
where $\sigma_{r}^2$ and $\mathbf{J}_{\xi}^p$ represent the variance of residual and its individual Jacobian vector of point $\mathbf{p}$ w.r.t. camera pose $\xi$ in Eq. (\ref{eq:edgeerror}). Therefore, the two components of search length representation can be upper bounded by enforcing the symmetric structure of search length as follows:\begin{subequations} 
\begin{gather}
\sigma_{p \perp g}  = \maxval \left( \sigma_1 \langle \mathbf{v}_1 , \mathbf{g}_{\perp} \rangle, \sigma_2 \langle \mathbf{v}_2 , \mathbf{g}_{\perp} \rangle \right) \label{eq:firstcomponent} \\
\sigma_{\mu}  = \maxval \left( \Vert \mathbf{p}(\xi,d \pm \sigma_{d})-\mathbf{p}(\xi,d) \Vert_2 \right) \label{eq:secondcomponent} 
\end{gather}
\end{subequations}
where $\mathbf{p}(\xi,d)$ denotes the point reprojection function described in Eq. (\ref{eq:pointproj}). The geometric interpretation of the derived search length centered at temporal correspondence ({\color{red} $\blacklozenge$}) is illustrated in Fig. \ref{fig:searcharea}.

\begin{figure}[t] 
  \centering
  \includegraphics[width=\linewidth]{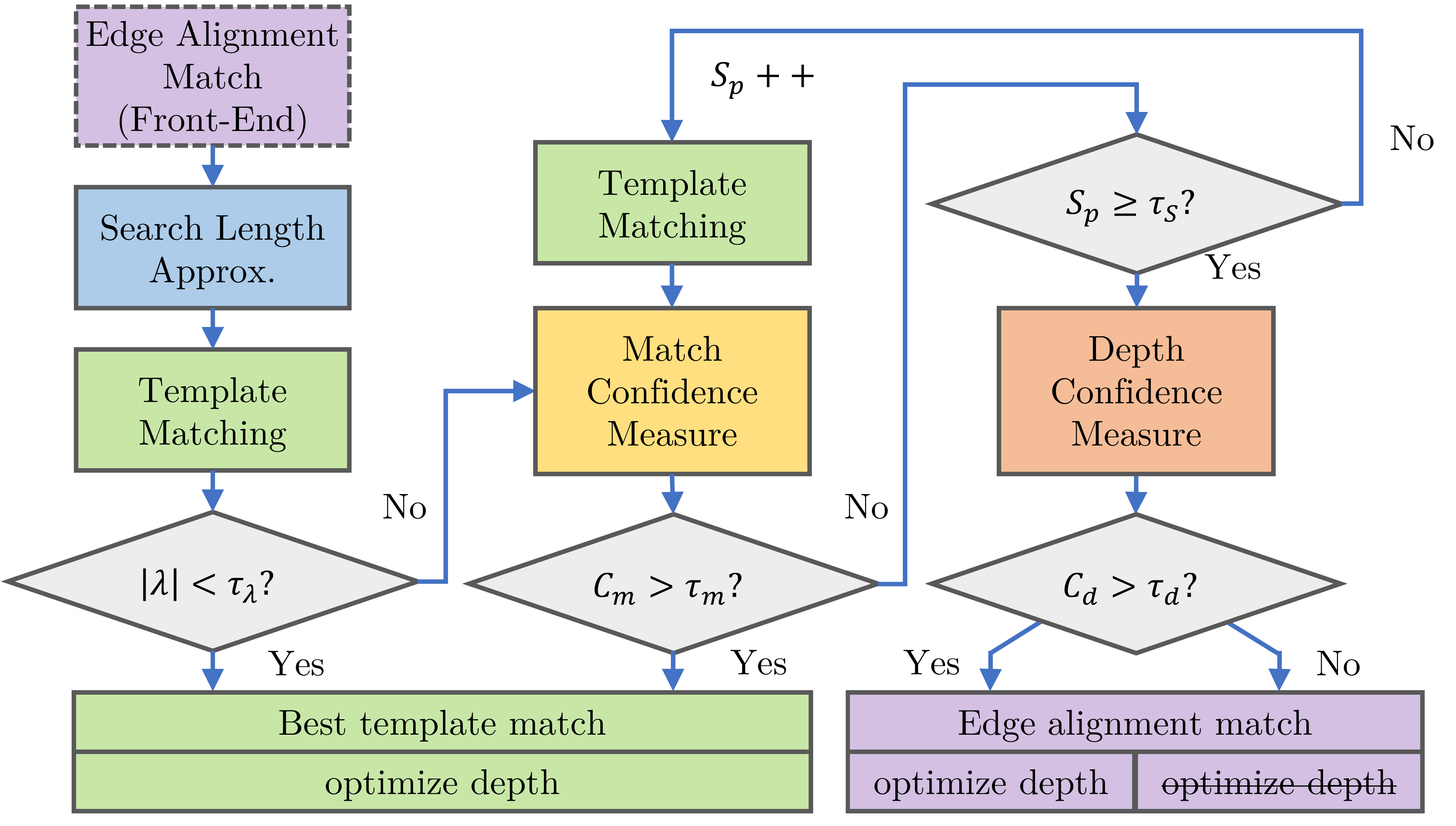}
    \caption{ \small \textbf{Edge-Guided Data Association Pipeline} incorporates probabilistic search length approximation (Sec. \ref{sssec:slapprox}), illumination-robust template matching (Sec. \ref{sssec:templatematching}) with match confidence based path size adaption (Sec. \ref{sssec:adaption}), and depth confidence based conditioning (Sec. \ref{sssec:depthconfidence}). Note that the edge alignment match comes directly from the proposed front-end, which doesn't involve any calculation here.}
\label{fig:pipeline}
\end{figure}

\subsubsection{Depth uncertainty}
\label{ssec:depthuncertainty}
Eliminating the search length factor $\lambda$ from Eq. \ref{eq:intersection}, the disparity estimate $\mu$ and its variance $\sigma_{\mu}^2$ can be expressed as:
\begin{subequations}
\begin{gather}
 \mu( \mathbf{p}) = \frac{\langle \mathbf{p}^* - \mathbf{p},  \mathbf{g} \rangle}{\langle\mathbf{g},\mathbf{l} \rangle} = \frac{e_{p \| g}}{cos \theta}  \label{eq:depth} \\
\sigma_{\mu}^2 = \mathbf{J}_{p} \Sigma_{p} \mathbf{J}_{p}^T =  \frac{\sigma_{p \| g}^2}{\cos^2 \theta} \label{eq:depthuncertainty} 
\end{gather}
\end{subequations}
where $\mathbf{J}_{p}$ is the Jacobian of Eqn. \ref{eq:depth} given the statistics of $\mathbf{p}$. $e_{p \| g}$ is the reprojection error component that is parallel to edge normal direction $\mathbf{g}$. Similar as the derivation in Eq. (\ref{eq:firstcomponent}), the $\sigma_{p \| g}$ can be expressed as follows:
\begin{equation} \label{eq:depthlength} 
\sigma_{p \| g}  = \maxval \left( \sigma_1 \langle \mathbf{v}_1 , \mathbf{g} \rangle, \sigma_2 \langle \mathbf{v}_2 , \mathbf{g} \rangle \right) 
\end{equation}

\subsection{Edge-Guided Data Association Pipeline}
\label{ssec:dapipeline}
The heart of our solution involves the proposed edge-guided data association pipeline, whose processing flow is depicted in Figure \ref{fig:pipeline} with input of edge alignment matches {\color{pink!40!blue!40!white} $\blacksquare$}. The important components include search length approximation {\color{blue!50!white} $\blacksquare$}, illumination-robust template matching {\color{green!50!white} $\blacksquare$}, match-confidence-based patch size adaption {\color{yellow!70!red} $\blacksquare$}, and depth confidence based match conditioning {\color{yellow!30!red} $\blacksquare$}.  

\subsubsection{Search length approximation}
\label{sssec:slapprox}
Given the camera transformation from the edge alignment front-end, all edge points can be projected onto newly added keyframe. Their nearest neighbors in a new keyframe then serve as the coarse initialization of edge point correspondence for further refinement in Fig. \ref{fig:ambiguity} (1). Then, the search length $\lambda$ is calculated for each point of interest using Eq. (\ref{eq:searchregion}) based on their statistics. 

\subsubsection{Illumination-robust template matching}
\label{sssec:templatematching}
Given the estimated search radius, our proposed probabilistic 1D search strategy starts with the coarsely estimated edge point correspondence. A standard region growing algorithm is used to explore the nearby edge points for template matching. To compensate for the rotation and scale difference between the matched patches, the patch is pre-transformed in Fig. \ref{fig:ambiguity} (2) based on estimated parameters $\mathbf{R}$, $\mathbf{t}$, and $\mathbf{d}$ by making the assumption that the transformation is locally constant. 

At each iteration, the template-based matching is performed through estimating the image gradient magnitude difference of a 5x5 patch ($S_p = 5$) between the query and the target edge points. The search stops at either maximum growth boundary or discontinuity of edges, where the edge point generating the smallest error is chosen as the best match for local bundle adjustment in Fig. \ref{fig:ambiguity} (3). 

\subsubsection{Match confidence based path size adaption}
\label{sssec:adaption}

\begin{figure}[t] 
  \centering
  \includegraphics[width=\linewidth]{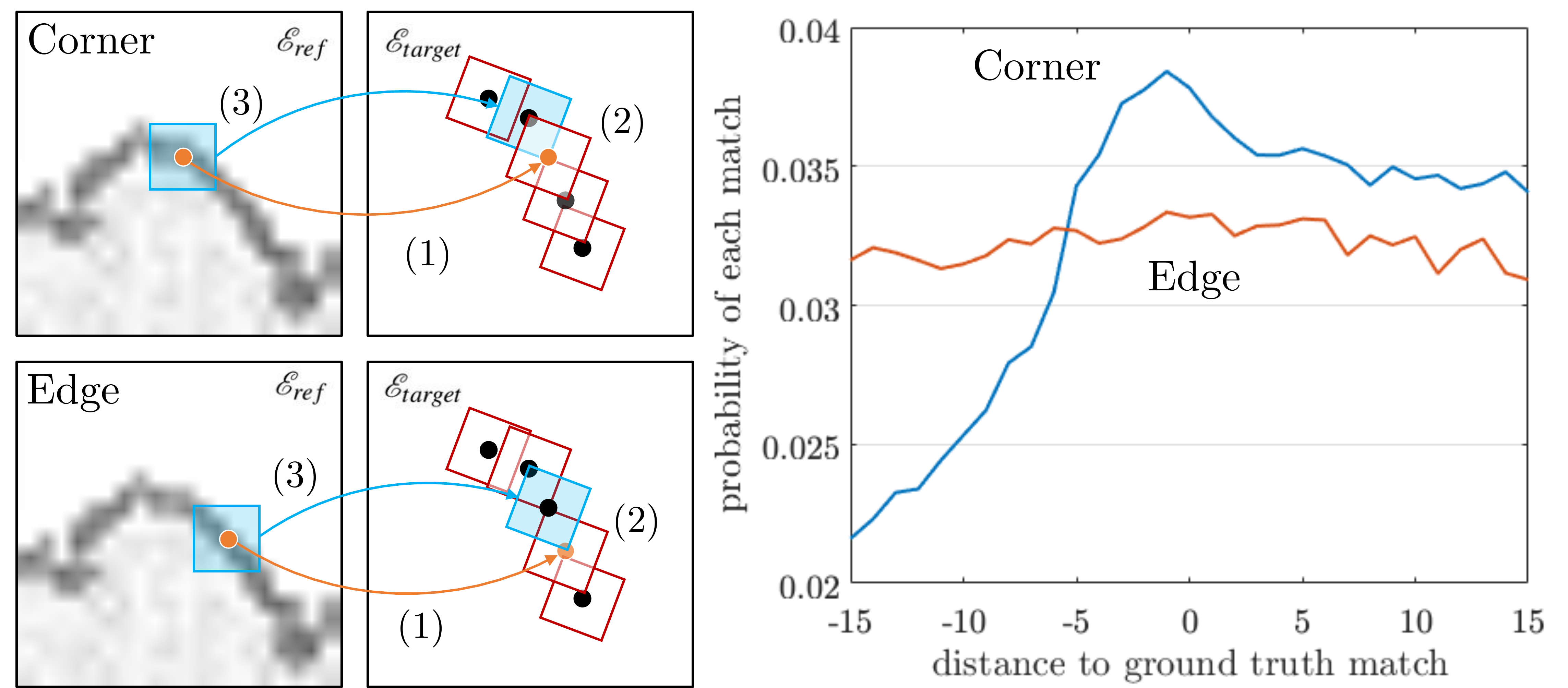}
    \caption{ \small \textbf{Our proposed template matching strategy and the potential ambiguity} is illustrated, where the general steps to realize the best match are labeled (left), while their match confidence measures (right), named Attainable Maximization Likelihood, are plotted to visualize the potential ambiguity for edge cases.}
\label{fig:ambiguity}
\end{figure}

The major drawback of the described naive template matching strategy is its ignorance of the potential ambiguity between locally similar edge points. It is especially true for flat edge cases in Fig. \ref{fig:ambiguity} (right), where a small pose error may generate large matching bias. Inspired by the research on confidence measures for stereo vision, the best-performance metric in \cite{hu2012quantitative}, named Attainable Maximum Likelihood (AML), is implemented to distinguish the ambiguous matches along edges as follows:
\begin{equation} \label{eq:amlconfidence} 
C_{m} = C_{AML} = \frac{1}{\sum_{\lambda} \Vert c_{\lambda} - c^* \Vert_2^2} 
\end{equation}
where $c_{\lambda}$ denotes the template matching cost within the search region, and $c^*$ means the cost of the best match. 

As long as the $C_{m}$ is smaller than a pre-defined threshold $\tau_{m}$, the patch size $S_p$ is increased until the patch size limit $\tau_S$ is met. The adaption of path size allows the algorithm to involve more information for template matching, which improves not only the noise resistance of patch matching \cite{engel2017direct} but also the accuracy of confidence measure \cite{brandao2015stereo}. 

Noted that we won't perform any match confidence check for points with very small search lengths, where a search length threshold $\tau_{\lambda}$ is pre-defined. A small search radius implies an accurate pose estimate, so that the best match within this region is also most likely to be the global optima. As a result, we decide to trust the naive matching approach to save computational resources. 

\subsubsection{Depth confidence conditioning}
\label{sssec:depthconfidence}
After the path size adaption process, the matches that still present high ambiguity ($C_{m} < \tau_{m}$) will be discarded. Instead of best photometric matches, the edge alignment matches will be passed into the later optimization framework and the depth confidence measure $C_d$ is proposed to predict the their observability at later depth estimation as follows:
\begin{equation} \label{eq:depthconfidence} 
C_d = \frac{1}{\sigma_{\mu}} =  \frac{cos \theta}{\sigma_{p \| g}}
\end{equation}
A small $C_d$ indicates a large potential depth estimation error, which means this particular match is unsuitable for depth estimation, and vice verse. The matches with low match $C_m$ and depth $C_d$ confidences will be passed into the local BA layer with a fixed depth at each iteration of joint optimization. 

\subsection{Data Association Updates}
\label{ssec:daupdate}
Inter-frame poses and scene structure can change significantly during local BA optimization so that the matches need to be updated during optimization. To capture the potential erroneous matches, we monitor the reprojection error component $e_{p \perp g}$, easily calculated from the local BA residuals $\mathbf{r}$. The update conditions can be expressed as follows:
\begin{equation} \label{eq:distancecheck} 
e_{p \perp g} = \langle \mathbf{r} , \mathbf{g}_{\perp} \rangle > k_m \lambda_{max}
\end{equation}
where $k_m$ is the ratio for match update that is typically set to be 0.5 -1.0. It means if the match distance on the edge direction is larger than a certain ratio of maximum search length, the match is most likely to be invalid and needs to be re-associated for estimation accuracy.  

\section{EVALUATION}
\label{sec:evaluation}
In this section, we evaluate our proposed monocular edge VO system quantitatively on publicly available datasets \cite{gaidon2016virtual}  \cite{griffith2017symphony} \cite{Geiger2012CVPR} using real-time capable edge detectors \cite{canny1987computational} \cite{dollar2013structured} \cite{xie2015holistically}. Compared to other edge VO systems concentrating on indoor navigation, our evaluation mainly focuses on challenging outdoor environments, where the sun-glare and pixel over-exposure are the main factors of tracking failure. Our experiments are conducted using a regular laptop featuring an Intel I7 core for our proposed VO pipeline, where the edge detection is calculated on an Nvidia K20 GPU.

\begin{figure}[t] 
  \centering
  \includegraphics[width=\linewidth]{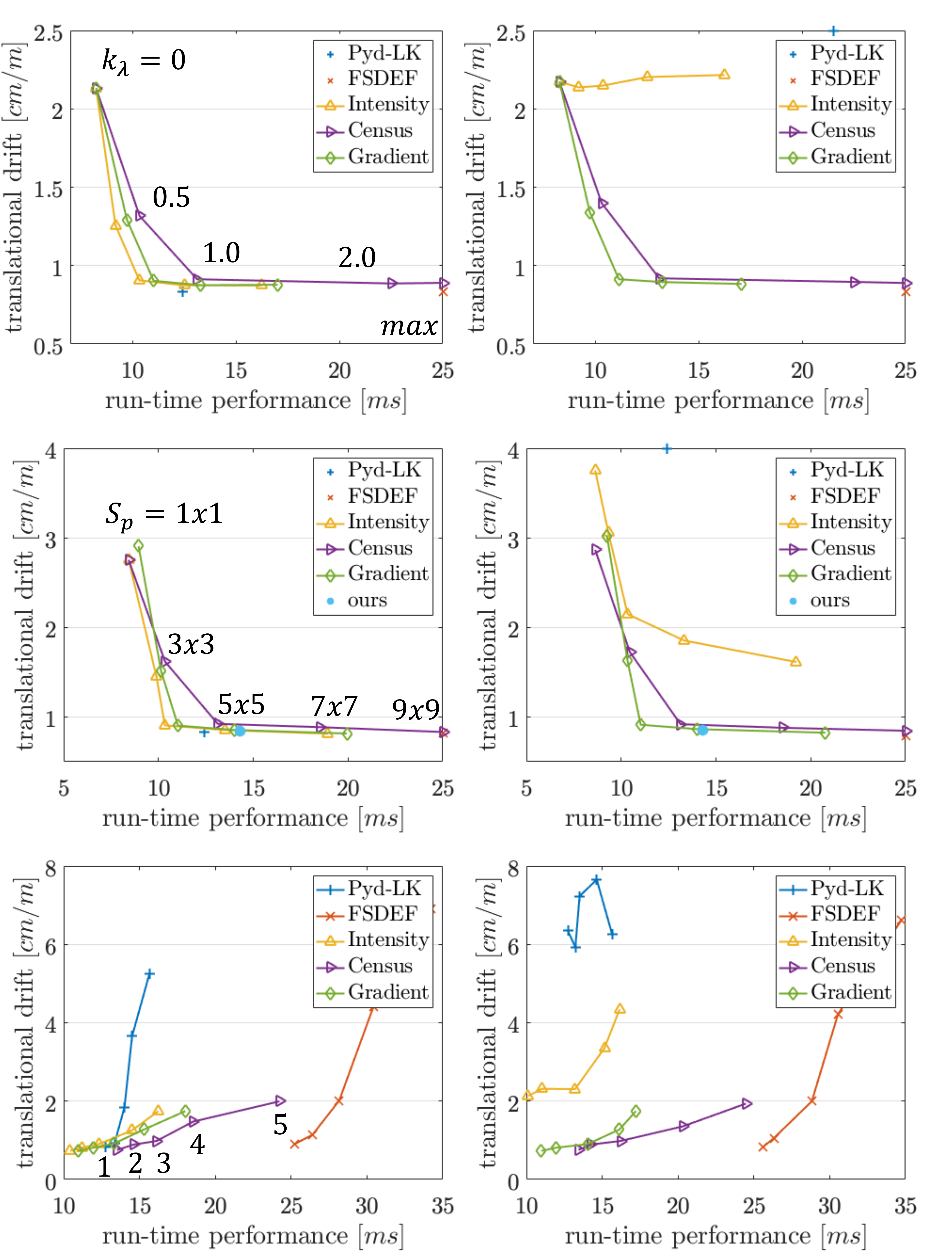}
    \caption{ \small \textbf{Our Proposed Edge-Guided Data Association} is evaluated under regular (left) and illumination-changing (right) sequences. The camera tracking accuracy and run-time performance are evaluated against (1) search length ratio $k_{\lambda} = \{ k_{p}, k_{d}\}$ in Eq. (\ref{eq:searchradius}) (top), (2) patch size $S_p$ (middle), and (3) inter-frame distances (bottom). It worth noting that the cases that $k_{\lambda} = 0$ and $k_{\lambda} = max$ are equivalent to the direct use of edge alignment tracking results and search with a fixed length, respectively.} 
\label{fig:paramstudy}
\end{figure}

\subsection{Evaluation on Edge-Guided Data Association }
\label{ssec:evaldataassociation}
First, we evaluate the accuracy and efficiency of our proposed edge-guided data association strategy against (1) search length, (2) patch size, and (3) inter-frame distance. The optical flow methods, e.g. the conventional Py-LK  \cite{baker2004lucas} and illumination-robust FSDEF \cite{garrigues2017fast}, serve as the baseline approaches for comparison. Intensity, census, and gradient approaches to template matching are plugged into our framework for evaluation, where the mean translational drift and average processing time are set as metrics for quantitative analysis. A photo-realistic vKITTI \cite{gaidon2016virtual} dataset with and without simulated illumination changes is used to evaluate our proposed data association approach, where the white Gaussian noise with $1.0$ standard variance is added to ground truth inverse depth. We uses 800 points uniformly sampled from edges in the image for evaluations.

The selection of translational drift as the evaluation metric, instead of the more direct average end-point error, is because of the difficulty to generate ground-truth matches for general edge points with sub-pixel precision and the lack of datasets that have the multi-view dense ground-truth pixel correspondences. Besides, our proposed system is designed to improve the camera tracking accuracy with real-time constraints, which motivates us to carry out \textit{end-to-end} evaluation.

In Fig. \ref{fig:paramstudy}, we can summarize some observations: (1) The \textit{gradient} approach shows the best overall performance, in terms of tracking accuracy and efficiency, in both regular and illumination-challenging environments. (2) Our proposed search radius formula with k = 1 works well on the tests, which is previously considered as underestimated due to the independence assumptions and multiple approximations. (3) Our proposed patch size adaption strategy allows consuming less time to realize better correspondences compared with fix-length search strategies. (4) Our proposed edge-guided data association strategy holds the better capability of dealing with large camera motions compared with optical flow methods.

\subsection{Evaluation on Illumination and Motion Robustness }
\label{ssec:evalillumination}

\begin{figure*}[t] 
  \centering
  \includegraphics[width=\linewidth]{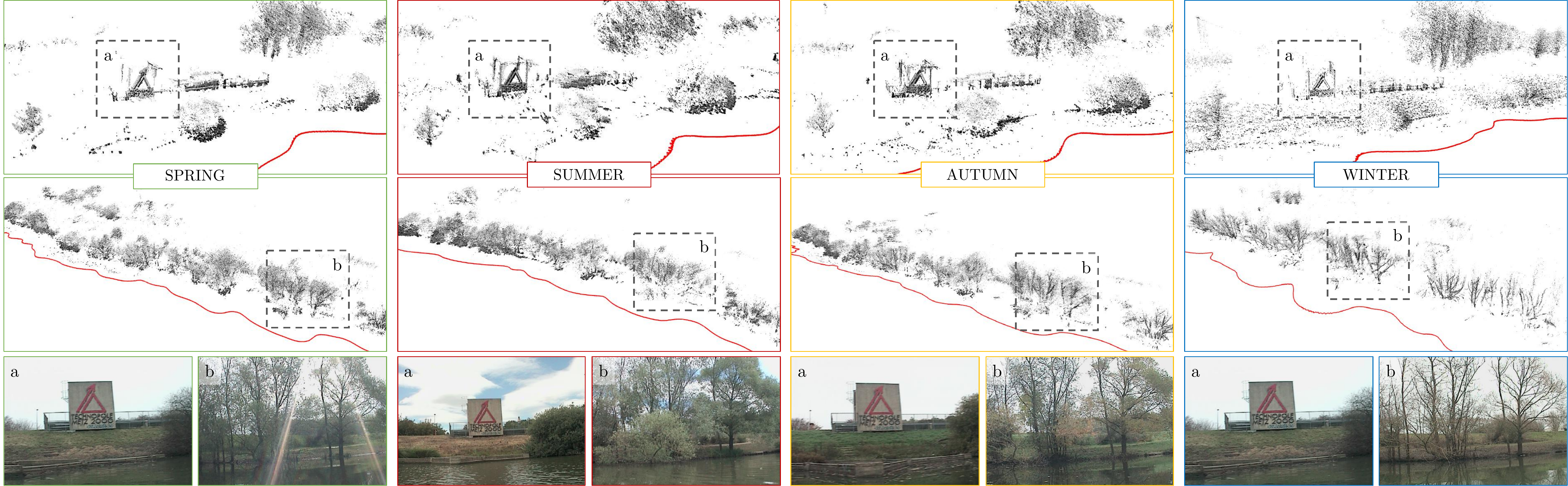}
    \caption{ \small \textbf{Evaluation on Symphony Lake Dataset}. The reconstructed pointclouds and example images of a human-made structure (top) and a natural scene (middle) are presented based on their seasons so that the structural and appearance changes across seasons can be observed. Although we merely present one sun-glare image at SPRING \textit{b}, sun-glares exists in the most test sequences.  }
\label{fig:reconstruction}
\end{figure*}

\begin{table*}[]
\centering
\caption{ System Evaluation against Illumination Changes and Fast Camera Motion.}
\resizebox{\textwidth}{!}{
\begin{tabular}{c|cccccccc|cccccccc|cc}
\hline
            & \multicolumn{8}{c|}{Symphony Lake Dataset}                                                                                                & \multicolumn{8}{c|}{Down-Sampled Symphony Lake Dataset }   &  \multicolumn{2}{c}{time}  \\
            & \multicolumn{2}{c}{\emph{spring}} & \multicolumn{2}{c}{\emph{summer}} & \multicolumn{2}{c}{\emph{autumn}} & \multicolumn{2}{c|}{\emph{winter}} & \multicolumn{2}{c}{\emph{spring}} & \multicolumn{2}{c}{\emph{summer}} & \multicolumn{2}{c}{\emph{autumn}} & \multicolumn{2}{c|}{\emph{winter}} & \multicolumn{2}{c}{ }            \\ 
            & rate          & err         & rate          & err          & rate          & err         & rate           & err           & rate          & err          & rate          & err       & rate    & err   & rate   & error   &     track & map                  \\ \hline
DSO\cite{engel2017direct}                               &  14.2  & 17.4 & 9.3    &  14.7  & 2.5  & 7.7   & 15.9  & 21.7  & 19.9 & 21.0 & 12.7  & 17.8 & 7.2 & 11.1 & 19.4 & 25.7 & 58  & 93\\
ORB\cite{mur2017orb}                                      &  23.5 & 26.3 & 16.4  &  19.3  & 8.8  & 13.3 & 19.2  & 24.6 & 24.1 & 26.6 & 16.0 & 22.1 & 8.9 & 13.5 & 19.6 & 24.8 & - & -\\
Prev\cite{wu2019illumination}                        &  4.7   & 9.7  & 4.4    &  9.2   &  \textbf{2.2}  & 6.4  & 8.0  & 12.4   & 10.2 & 14.7  & 7.6   & 11.5   & 7.3 & 11.4 & 14.1 &  18.5& 105 & 173\\ \hline 
Ours(Canny\cite{canny1987computational}) 	&  3.3   & 8.1    & 4.9    &  9.6  &  \textbf{2.2}  & 6.4  &  \textbf{2.4}   &  6.3   & 3.6  &  8.9  & 5.1    & 10.8   & 3.4 & 8.2 & 2.5 &  6.4& 77 & 201\\ 
Ours(SE\cite{dollar2013structured}) 				&  \textbf{1.8}    & \textbf{6.8}   & 3.3    &  8.4   &  \textbf{2.2}  & 6.4   &  \textbf{2.4}   &  \textbf{6.1}   &  \textbf{2.1}   &  \textbf{7.0}   & 4.9    &  \textbf{10.1}   & 2.3 & 6.6 &  \textbf{2.3} & \textbf{6.2} & 89 & 193\\ 
Ours(HED\cite{xie2015holistically}) 				&  1.9    & 6.9    &  \textbf{3.1}    &   \textbf{8.2}   &  \textbf{2.2}  &  \textbf{6.3}   &  \textbf{2.4}   &   \textbf{6.1}   & 2.2   & 7.1  &  \textbf{4.8}    & \textbf{10.1}    &  \textbf{2.2} &  \textbf{6.4} &  2.4 &  \textbf{6.2} &  92 & 194\\ \hline
\end{tabular}
}
\vspace{1mm}
Rates are failure rate per sequence, averaged over 10 trials. Err is a drift rate in [$cm/m$]. Processing time in [$ms$] per frame. 

\label{table:evalrobustness}
\end{table*}
 
\begin{table}[htb]
\centering
\caption{ System Evaluation on Normal Sequences.}
\resizebox{0.45\textwidth}{!}{
\begin{tabular}{|c|ccccc|}
\hline
KITTI & DSO                                  & ORB                           & Ours(Canny)                                    &  Ours(SE)                                 & Ours(HED)           \\ 
\cite{Geiger2012CVPR}	     & \cite{engel2017direct}     & \cite{mur2017orb}    &  \cite{canny1987computational}    & \cite{dollar2013structured}   & \cite{xie2015holistically}           \\   \hline
00        & 16.83    & 16.14                 & 18.36  & \textbf{16.06} & 16.11 \\
01        & 36.32    & -                        & 32.59  & 22.31               & \textbf{21.55} \\
02        & 17.08    & \textbf{15.58}   & 17.72   & 16.77               & 16.52          \\
03        & 3.71      & \textbf{3.44}     & 4.31    & 3.64                 & 3.63           \\
04        & 3.01      & 3.05                   & 2.93    & 2.41                  & \textbf{2.33} \\
05        & 13.64   & \textbf{12.96}    & 13.49  & 13.05               & 13.02           \\
06        & 14.13    & 13.35                  & 13.44  & \textbf{12.54} & 12.57           \\
07        & 9.55     & 9.63                  & 11.36    & \textbf{8.15}   & 8.20           \\
08        & 18.31   & \textbf{15.43}    & 19.35   & 16.24               & 16.21          \\
09        & 13.05   & 12.88                  & 12.73   & 12.61                & \textbf{12.47} \\ \hline
Avg.(*) &12.15    & 11.50                   & 12.63  & 11.52                &  \textbf{11.44} \\ \hline
\end{tabular}         
}
\vspace{1mm} \\
$(*)$ excludes Kitti 01 sequence.  
\label{table:evalKITTI}
\end{table}

The overall system performance of our proposed edge VO algorithm is evaluated using Symphony Lake \cite{griffith2017symphony} dataset, which consists of millions of natural lakeshore images heavily contaminated by (1) smooth (auto-exposure) and (2) sudden (sun-glares) lighting changes, as well as (3) the tree-sky boundary pixel over-exposure. Unlike sun-glares that randomly occurs in sunny days of a year, the over-exposure induced appearance changes show significant variations across seasons. In general, the denser the leaves in Summer, the fewer boundary pixels are 'eaten' by lights the VO system is, therefore, less affected by over-exposure, and the opposite holds in Winter. Based on these observations, we choose 12 surveys (3 surveys per season) heavily contaminated with sun-glare and categorize results based on seasons. For motion robustness evaluation, we down-sample the selected sequences at a sample rate of 3 for fast camera motion simulation. For quantitative analysis, the ground truth pose is calculated through Laser-GPS-based global pose graph optimization described in \cite{pradalier2018multi}. The scale of trajectory from monocular VO are corrected using ground truth poses at every 200 frames, while the loop-closure functionality is disabled for ORBSLAM2 for a fair comparison.

In Fig. \ref{fig:reconstruction}, the pointclouds show that our proposed approach is capable of reconstructing high-quality human-made structures as well as the natural scenes under illumination contamination. Comparing reconstruction results across seasons, we can observe the structural changes of trees between different times of the year. A more detailed quantitative evaluation concerning relative pose errors (RPEs), failure rate, and runtime performance using complete and down-sampled sequences are presented in Table. \ref{table:evalrobustness}. The camera pose error within tracking failure locations is calculated using pre-defined maximum tracking error (30.0 $cm/m$), where the large errors indicate both low tracking accuracy and low completion rate.

In general, our proposed approach outperforms all other state-of-the-art methods, where the \textit{Winter} sequences even show better performance than the \textit{Summer} ones. Such variation can be attributed to the fact that the \textit{Winter} images generally hold more clear and uniformly distributed edges compared with those of \textit{Summer}. Among different edge detectors, the learned edges (SE \cite{dollar2013structured}, HED \cite{xie2015holistically}) present better tracking accuracy and robustness over conventional Canny edge detector, most likely due to the low repeatability of Canny edges in the outdoor image sequences \cite{wu2019semantic}. 

For the evaluation on the influence of high-speed camera motions, it can be observed that both the failure rates and tracking error of state-of-art methods increase significantly after frame sub-sampling in comparison of our proposed approach. Similar to our previous analysis, the \textit{Winter} sequences using learned edges shows the least performance degradation compared with other approaches. It suggests that well-distributed high-repeatability edges are essential for our proposed edge VO framework, which guarantees both the robustness against illumination changes as well as high-speed camera motions.  

\subsection{Evaluation on Regular Sequences }
\label{ssec:evalmotion}
Besides challenging illumination- and motion-challenging dataset, we also evaluate our proposed system on regular sequences for completeness. Table. \ref{table:evalKITTI} shows the absolute trajectory errors (ATEs) after scale correction using ground truth poses on KITTI \cite{Geiger2012CVPR} dataset. Our proposed edge VO approach shows consistent improvements over direct DSO and achieves comparable performance with indirect ORBSLAM2.

\section{CONCLUSIONS}
In this work, we propose a monocular edge visual odometry framework, which is a real-time capable algorithm exploiting the edge features and image gradient for illumination-robust camera motion estimation and scene reconstruction. These are obtained by an edge alignment front-end, a finer point correspondence refinement strategy through a fast probabilistic 1D search strategy, and joint optimization in local bundle adjustment. The proposed system successfully overcomes the partial observability issue of monocular edge mapping as well as improving the robustness of outdoor motion estimation. The experimental results indicate that our proposed system outperforms current state-of-art algorithms in terms of illumination- and motion-robustness and shows comparable performance in regular sequences. 

%%%%%%%%%%%%%%%%%%%%%%%%%%%%%%%%%%%%%%%%%%%%%%%%%%%%%%%%%%%%%%%%%%%%%%%%%%%%%%%%

\addtolength{\textheight}{-0cm}   % This command serves to balance the column lengths

%%%%%%%%%%%%%%%%%%%%%%%%%%%%%%%%%%%%%%%%%%%%%%%%%%%%%%%%%%%%%%%%%%%%%%%%%%%%%%%%

%%%%%%%%%%%%%%%%%%%%%%%%%%%%%%%%%%%%%%%%%%%%%%%%%%%%%%%%%%%%%%%%%%%%%%%%%%%%%%%%

%%%%%%%%%%%%%%%%%%%%%%%%%%%%%%%%%%%%%%%%%%%%%%%%%%%%%%%%%%%%%%%%%%%%%%%%%%%%%%%%

%%%%%%%%%%%%%%%%%%%%%%%%%%%%%%%%%%%%%%%%%%%%%%%%%%%%%%%%%%%%%%%%%%%%%%%%%%%%%%%%

{\small
\bibliographystyle{IEEEtran}
\bibliography{IEEEabrv,root}
}

\end{document}